\documentclass[letterpaper]{article}
\usepackage{iccc}

\usepackage{graphicx}
\usepackage[fleqn]{amsmath}
\usepackage{amssymb}
\usepackage{times}
\usepackage{helvet}
\usepackage{courier}
\usepackage{booktabs}
\usepackage{multirow}
\usepackage{float}
\usepackage[bordercolor=white,backgroundcolor=gray!30,linecolor=black,colorinlistoftodos]{todonotes}

\title{A Pig, an Angel and a Cactus Walk Into a Blender: \\ 
A Descriptive Approach to Visual Blending\\
}
\author{Jo\~{a}o M. Cunha, Jo\~{a}o Gon\c{c}alves, Pedro Martins, Penousal Machado, Am\'{i}lcar Cardoso\\
CISUC, Department of Informatics Engineering\\
University of Coimbra\\
\{jmacunha,jcgonc,pjmm,machado,amilcar\}@dei.uc.pt\\
}
\begin{document} 
\maketitle
\begin{abstract}

A descriptive approach for automatic generation of visual blends is presented. The implemented system, the \emph{Blender}, is composed of two components: the \emph{Mapper} and the \emph{Visual Blender}. The approach uses structured visual representations along with sets of visual relations which describe how the elements -- in which the visual representation can be decomposed -- relate among each other. Our system is a hybrid blender, as the blending process starts at the \emph{Mapper} (conceptual level) and ends at the \emph{Visual Blender} (visual representation level). The experimental results show that the \emph{Blender} is able to create analogies from input mental spaces and produce well-composed blends, which follow the rules imposed by its base-analogy and its relations. The resulting blends are visually interesting and some can be considered as unexpected. 

\end{abstract}

\textcolor{red}{Cite as: Cunha, J. M.; Gon\c{c}alves, J.; Martins, P.; Machado, P.; and Cardoso, A. 2017. A pig, an angel and a cactus walk into a blender: A descriptive approach to visual blending. In: Proceedings of the Eighth International Conference on Computational Creativity (ICCC 2017).}

\section{Introduction}
Conceptual Blending (CB) theory is a cognitive framework proposed by Fauconnier and Turner \shortcite{fauconnier2002} as an attempt to explain the creation of meaning and insight. CB consists in integrating two or more mental spaces in order to produce a new one, the blend(ed) space. Here, mental space means a temporary knowledge structure created for the purpose of local understanding~\cite{Fauconnier:94}.

Visual blending, which draws inspiration from CB theory, is a relatively common technique used in Computational Creativity to generate creative artefacts in the visual domain. While some of the works are explicitly based on Conceptual Blending theory, as blending occurs at a conceptual level, other approaches generate blends only at a representation/instance level by means of, for example, image processing techniques.

We present a system for automatic generation of visual blends (\emph{Blender}), which is divided into two different parts: the \emph{Mapper} and the \emph{Visual Blender}.
We follow a descriptive approach in which a visual representation for a given concept is constructed as a well-structured object (from here onwards when we use the term representation we are referring to visual representations). The object can contain other objects and has a list of descriptive relations, which describe how the object relates to others. The relations describe how the representation is constructed (example: \emph{part A} inside \emph{part B}). In our opinion, this approach allows an easier blending process and contributes to the overall sense of cohesion among the parts.

%CAMERA READY JOAO: adicionei part GA

Our system can be seen as a hybrid blender, as the blending process starts at the conceptual level (which occurs in the \emph{Mapper}) and only ends at the visual representation level (which occurs in the \emph{Visual Blender}). We use an evolutionary engine based on a Genetic Algorithm, in which each population corresponds to a different analogy and each individual is a visual blend. The evolution is guided by a fitness function that assesses the quality of each blend based on the satisfied relations. In the scope of this work, the focus is given to the \emph{Visual Blender}.

\begin{figure}
\includegraphics[width=8.41 cm]{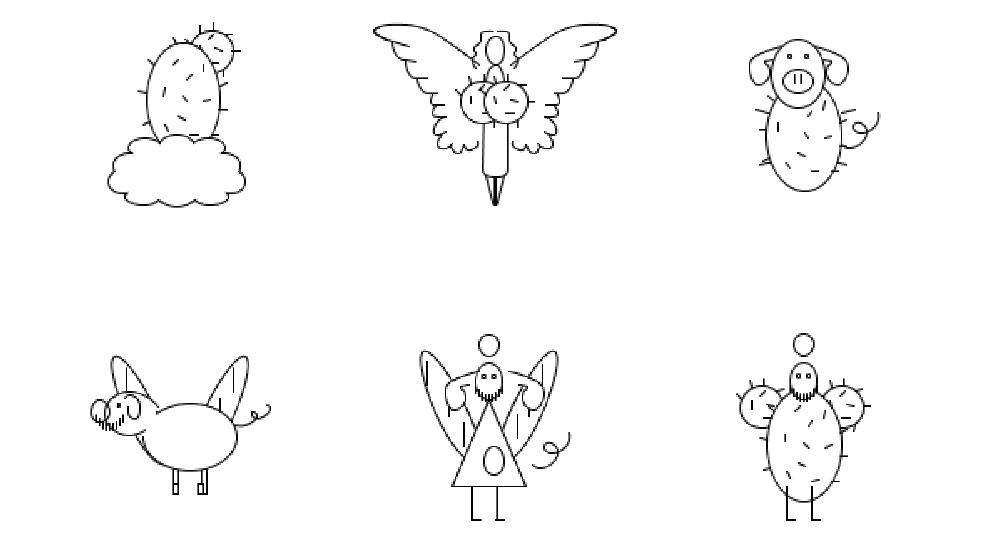}
\caption{Examples of produced blends.
}
\label{fig:blends1} %id para referenciar
\end{figure}

\section{Related Work}

In terms of the type of rendering, current computational approaches to visual blending can be divided into two groups: the ones which attempt to blend pictures or photorealistic renderings; and the ones that focus on non-photorealistic representations, such as pictograms or icons.

%Penousal: esta frase (mais propriamente o conteudo desta frase poderia passar para a intro, é uma distinção importante

The Boat-House Visual Blending Experience~\cite{pereira2002boat} is, to the best of our knowledge, one of the earliest attempts to computationally produce visual blends. The work was motivated by the need to interpret and visualize blends produced by a preliminary version of the Divago framework, which is one of the first artificial creative systems based on CB theory~\cite{pereira2007creativity}. In addition to a declarative description of the concepts via rules and concept maps (i.e., graphs representing binary relations between concepts), Pereira and Cardoso also considered a domain of instances, which were drawn using a Logo-like programming language. To test the system, the authors performed several experiments with the \emph{house} and \emph{boat} blend~\cite{goguen1999introduction} considering different instances for the input spaces.

%Penousal: besides não tem bem o significado que tu pensas que tem (não significa em adição/ para além)

Ribeiro et al. \shortcite{ribeiro2003} explored the use of the Divago framework in procedural content generation. In this work, the role of Divago was to produce novel creatures at a conceptual level from a set of existing ones. Then, a 3D interpreter was used to visualize the objects. The interpreter was able to convert concept maps from Divago, representing creatures, into Wavefront OBJ files that could be rendered afterwards.

%penousal: assumo que esta seja do tipo 2...
Steinbr\"uck \shortcite{steinbruck2013conceptual} introduced a framework that formalises the process of CB while applying it to the visual domain. The framework is composed of five modules that combine image processing techniques with gathering semantic knowledge about the concept depicted in an image with the help of ontologies. Elements of the image are replaced with other unexpected elements of similar shape (for example, round medical tablets are replaced with pictures of a globe).

Confalonieri et al. \shortcite{confalonieri2015} proposed a discursive approach to evaluate the quality of blends (although there is no evidence of an implementation). The main idea was to use Lakatosian argumentative dialogue~\cite{Lakatos76} to iteratively construct valuable and novel blends as opposed to a strictly combinatorial approach. To exemplify the argumentative  approach, the authors focused on  icon design by introducing a semiotic system for modelling computer icons. Since icons can be considered as a combination of signs that can convey multiple intended meanings to the icon, Confalonieri et al. proposed argumentation to evaluate and refine the quality of the icons. 

Xiao and Linkola \shortcite{Xiao2015} proposed Vismantic, a semi-automatic system aimed at producing visual compositions to express specific meanings, namely the ones of abstract concepts. Their system is based on three binary image operations (juxtaposition, replacement and fusion), which are the basic operations to represent visual metaphors~\cite{phillips2004beyond}. For example, Vismantic represents the slogan \emph{Electricity is green} as an image of an electric light bulb where the wire filament and screw base are fused with an image of green leaves. The selection of images as well as the application of the visual operations require user's intervention.

Correia et al. \shortcite{correia2016x} proposed X-Faces, which can be seen as a data augmentation technique to autonomously generate new faces out of existing ones. Elementary parts of the faces, such as eyes, nose or mouth, are recombined by means of evolutionary algorithms and computer vision techniques. The X-Faces framework generates unexpected, yet realistic, faces by exploring the shortcomings and vulnerabilities of computational face detectors to promote the evolution of faces that are not recognised as such by these systems.

Recent works  such as DeepStyle~\cite{gatys2015neural} can also be seen as a form of visual blending. DeepStyle is based on a deep neural network that has the ability to separate image content from certain aspects of style, allowing to recombine the content of an arbitrary image with a given rendering style (style transfer). The system is known for mimicking features of different painting styles.

Several other authors have seen the potential of deep neural networks for tasks related to visual blending~\cite{berov2016visual,mccaig2016deep,heath2016before}. For instance, Berov and  K\"{u}hnberger~\shortcite{berov2016visual} proposed a computational model of visual hallucination based on deep neural networks. To some extent, the creations of this system can be seen as visual blends.

%CAMERA READY JOAO: adicionar artigo google?

\section{The approach}

%Penousal: primeira frase tem que ser revista, não consigo perceber, deve haver aqui uns 4 conceitos distintos. Tem que se perceber a primeira frase de uma abordagem. Parte em pelo menos duas: a nossa inspiração é esta. depois o resto (que não consigo mesmo perceber)
Having the organization of mental spaces as an inspiration, we follow a similar approach to structure the construction of the visual representations, which are considered as a group of several parts / elements. 
%Penousal: traduzinfo, o que estas a dizer é que vais considerr que uma representação visual é constituida por um conjunto de elementos e relações entre eles (foi o que percebi)... esta manhosito
By focusing on the parts instead of the whole, there is something extra that stands out: not only is given importance to the parts but the representation ceases to be a whole and starts to be seen as parts related to each other.
%Penousal: ainda não acabaste de descrever a representação, e saltas para a justificação da representação... não fica muito mal
As our goal is to produce visual results, these relations have a visual descriptive nature (i.e. the nature of the relation between two elements is either related to their relative position or to their visual qualities). 
%Penousal: this allows the generation OU this allows us to generate... mas, mais importante, eu não sei qual é a base-representation... não esta dito...
This allows the generation of visual blends, guided and evaluated by criteria imposed by the relations present in the base-representations (see Fig.\ref{fig:repBase}) used in the visual blend production.

%% Penousal sugestão: tens que começar por dizer: os blends vão ser gerados usando este sistema assim e assado que usa uma representação deste género.
%% Depois dises os visual blends vão ser feitos assim e assado

%me alternativa dizes: queremos fazer blends visuais que tenham estas propriedades (basicamente as que estão descritas abaixo, depois, partindo dessa motivação dizes para o efeito bla bla bla (ou seja o que eu disse acima)...

%Penousal: esta está bem...
In addition, by using a representation style that consists of basic shapes, we reduce the concept to its simplest form, maintaining its most important features and thus, hopefully, capturing its essence (a similar process can be seen in Picasso's \emph{The Bull}, a set of eleven lithographs produced in 1945). As such, our approach can be classified as belonging to the group of non-photorealistic visual blending. This simplification of concepts has as inspiration several attempts to produce a universal language, understandable by everyone -- such as the pictographic ISOTYPE by Otto Neurath \shortcite{neurath1936international} or the symbolic Blissymbolics by Charles Bliss \shortcite{bliss1965semantography}.

%CAMERA READY JOAO: ISTO FOI MUDADO
As already mentioned, our main idea is centered on the fact that the construction of a visual representation for a given concept can be approached in a structured way. 
%The representation is therefore not only seen as a whole but as a compound of parts related to each other.
Each representation is associated with a list of descriptive relations (e.g.: \emph{part A} below \emph{part B}), which describes how the representation is constructed. Due to this, a visual blend between two representations is not simply a replacement of parts but its quality is assessed based on the number of relations that are respected. This gives much more flexibility to the construction of representations by presenting a version of it and also allowing the generation of similar ones, if needed. 

The initial idea involved only a representation for each concept. However, a given concept has several possible visual representations (e.g. there are several possible ways of visually representing the concept \emph{car}), which means that only using one would make the system very limited.

In order to avoid biased results, we decided to use several versions for each concept. Each visual representation can be different (varying in terms of style, complexity, number of characteristics and even chosen perspective) and thus also having a different set of visual relations among the parts. 

%CAMERA READY JOAO: comparação com os outros

In comparison to the systems described in the previous Section, we follow a different approach to the generation of visual blends by implementing a hybrid system and giving great importance to the parts and their relations -- such tends to be overlooked by the majority of the reviewed works in which an unguided replacement of parts often leads to a lack of cohesion among them. This approach allows us not only to assess the quality of the blends and guide evolution but also to easily generate similar (and also valid) blends based on a set of relations.

\subsection{Collecting data}
\label{se:colleting}

% the base-representations (how and why?)

%, which would be used to produce visual representations

The initial phase of the project consisted in a process of data collection. Firstly, a list of possible concepts was produced by collecting concepts already used in the conceptual blending field of research. 
%pnousal: que lista?
From this list, three concepts were selected based on their characteristics: \emph{angel} (human-like), \emph{pig} (animal) and \emph{cactus} (plant) -- collected from \citeauthor{Costello:CC_is_not_Structure_Alignment01} \shortcite{Costello:CC_is_not_Structure_Alignment01}. The goal of this phase was to collect visual representations for these concepts. An enquiry to collect the desired data was designed, which was composed of five tasks:
\begin{description}
\item [T1] Collection of visual representations for the selected concepts;
\item [T2] Identification of the representational elements;
%eu usaria o termo elementos em vez de parts, não mudei porque pode escapar alguma
\item [T3] Description of the relations among the identified elements;
\item [T4] Identification of the prototypical elements -- i.e. the element(s) that most identify a given concept \cite{johnson1985prototype}. For instance, for the concept \emph{pig} most participants considered \emph{nose} and \emph{tail} as the prototypical elements;
\item [T5] Collection of visual blends for the selected concepts.
\end{description}

\begin{figure}
\includegraphics[width=7.5 cm]{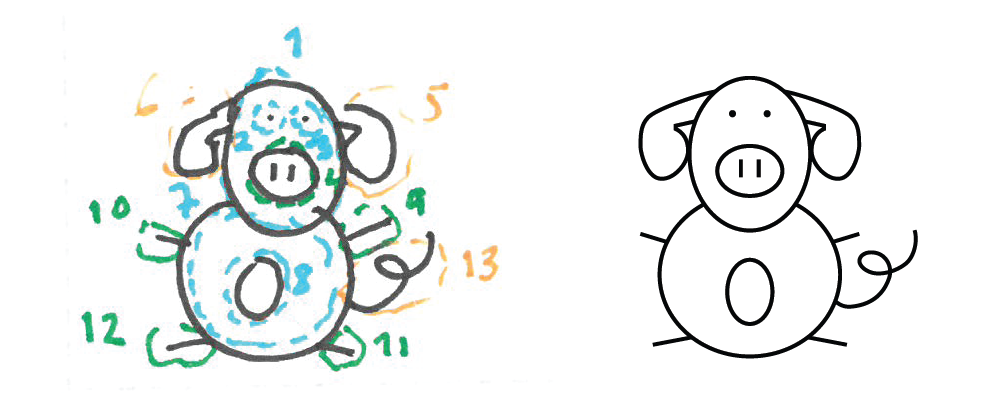}
\caption{On the left is a the representation drawn with the elements identified; On the right is the result of the conversion into fully scalable vector graphic.
}
\label{fig:repConvertion} %id para referenciar
\end{figure}

The data was collected from nine participants who were asked to complete the required tasks. In the first task (T1), the participants were asked to draw a representation for each concept avoiding unnecessary complexity but still representing the most important elements of the concept. In order to achieve intelligible and relatively simple representations, the participants were suggested to use primitives such as lines, ellipses, triangles and quadrilaterals as the basis for their drawings. After completing the first version, a second one was requested. The reason for two versions was to promote diversity. 

In the second task (T2), the participants identified the elements drawn using their own terms (for example, for the concept \emph{angel} some of the identified elements were \emph{head}, \emph{halo}, \emph{legs}). 

%CAMERA READY JOAO: Mudei isto
After completing the previous task, the participants were asked to identify the relations among elements that they considered as being essential essential (T3). These relations were not only related to the conceptual space but also (and mostly) to the representation. In order to help the participants, a list of relations was provided. Despite being told that the list was only to be considered as an example and not to be seen as closed, all the participants used the relations provided -- this ensured the semantic sharing between participants. Some participants suggested other relations that were not on the list --  these contributions were well-received.

\begin{figure}
\includegraphics[width=8.41 cm]{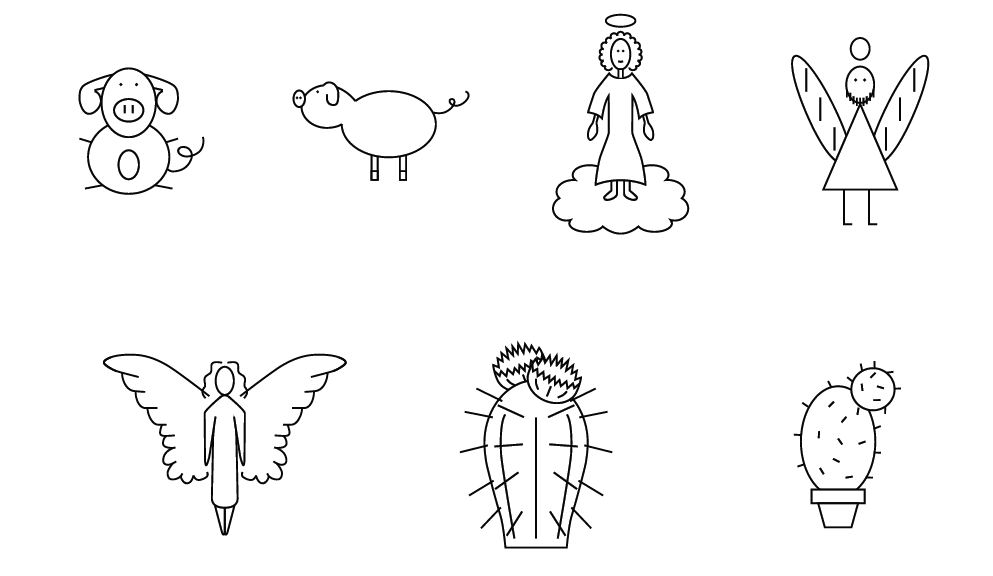}
\caption{Representations used as a base.}
\label{fig:repBase} %id para referenciar
\end{figure}

%CAMERA READY JOAO: Mudei isto
The identified relations are dependent on the author's interpretation of the concept, which can be divided into two levels. The first level is related to how the author interprets the connections among the concepts of the parts at a conceptual level (for example \emph{car}, \emph{wheel} or \emph{trunk}). The second level is related to the visual representation being considered: different visual representations may have different relations among the same parts (this can be caused, for example, by the change of perspective or style) -- e.g. the different positioning of the head in the two \emph{pig} representations in Fig.\ref{fig:repBase}.

Task four (T4) consisted in identifying the prototypical parts of the representations -- the parts which most identify the concept \cite{johnson1985prototype}. These will be used for interpreting the results obtained and for posterior developments.
%most e não mostly (maioria versus maioritariamente )

In the last task of the enquiry (T5), the participants were asked to draw representations for the blends between the three concepts. As a blend between two concepts can be interpreted and posteriorly represented in different ways (e.g. just at a naming level a blend between \emph{pig} and \emph{cactus} can be differently interpreted depending on its name being \emph{pig-cactus} or \emph{cactus-pig}). For this reason, the participants were asked to draw one or more visual representations for the blend. These visual representations were later used for comparing with the results obtained with the \emph{Visual Blender}.
%penousal: aqui ja tens visual representation

\subsection{Post-enquiry}
After the conduction of the enquiry, the data was treated in order to be used by the \emph{Visual Blender}. Firstly, the representations collected for each of the concepts were converted into fully scalable vector graphics (see Fig. \ref{fig:repConvertion}) and prepared to be used as base visual representations (see Fig.\ref{fig:repBase}) for the \emph{Visual Blender} (using layer naming according to the data collected for each representation -- each layer was named after its identified part). In addition to this, the relations among parts were formatted to be used as input together with their corresponding representation.
%penousal: não sei se se percebe o que é um layer (eu percebi)

\begin{figure}
\includegraphics[width=\linewidth]{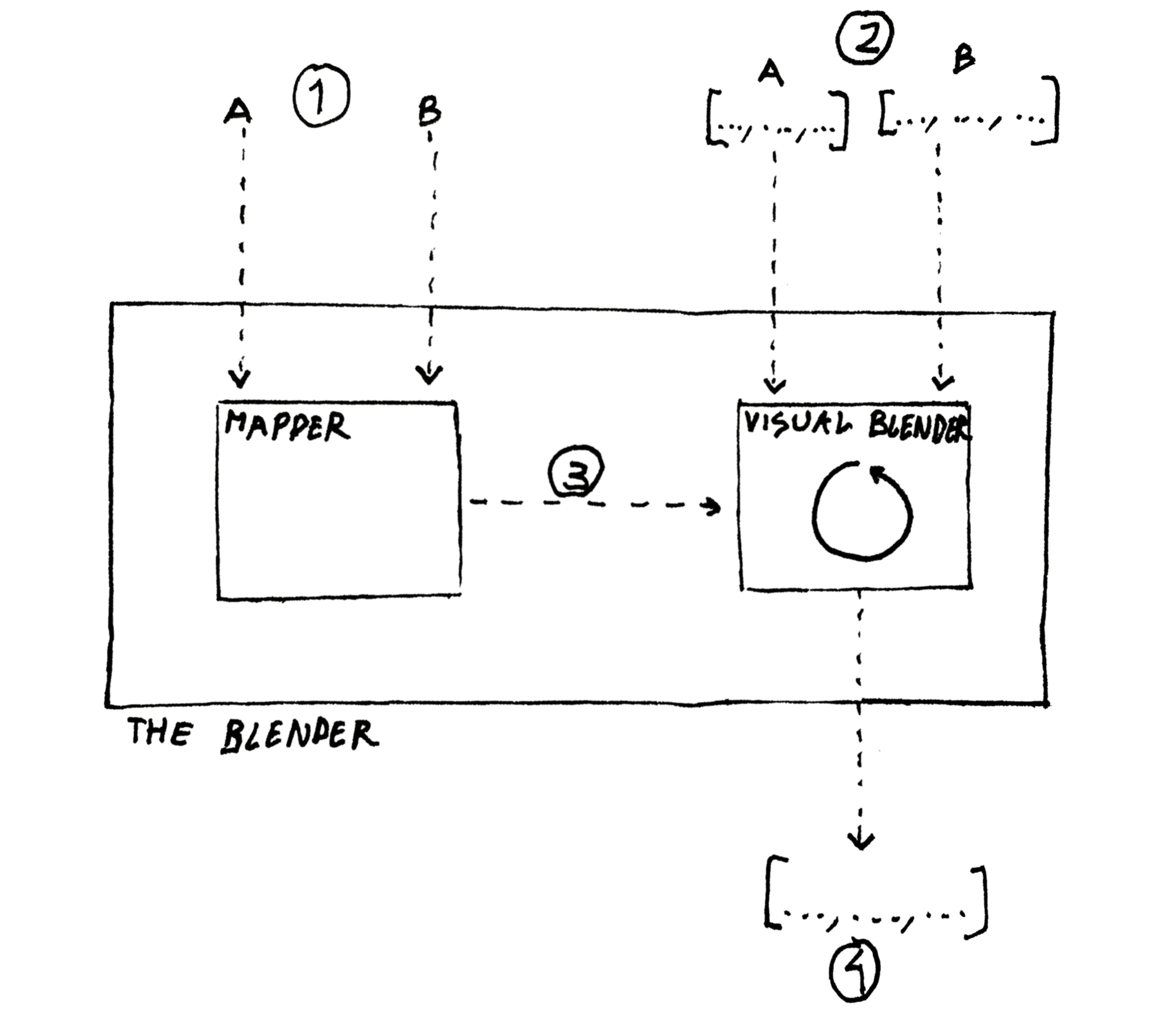}
\caption{Structure of the implemented Blender. The Blender consists of a Mapper and a visual Blender. The figure also shows the input spaces (1), the visual representations and list of relations (2), the produced analogies (3) and the produced blends (4).
}
\label{fig:structure} %id para referenciar
\end{figure}

\section{The Visual Blender}

As already mentioned, the \emph{Blender} has two different components: the \emph{Mapper} and the \emph{Visual Blender} (see Fig.\ref{fig:structure}). The \emph{Mapper} receives two input spaces (represented as 1 in Fig.\ref{fig:structure}), one referring to \emph{concept A} and the other one to \emph{concept B}. It produces analogies (3 in Fig.\ref{fig:structure}) that are afterwards used by the \emph{Visual Blender} component. The \emph{Visual Blender} also receives visual representations and corresponding list of relations among parts (2 in Fig.\ref{fig:structure}) that are used as a base and data for producing the visual blends (4 in Fig.\ref{fig:structure}).

%Camera Ready JOAO: adicionei isto
As this paper is focused on the \emph{Visual Blender} component, the Mapper is only briefly described (subsection \emph{Generating the blends: structural mapping}). Despite being related, the two components have different implementation details (e.g. object structure).

\subsection{Generating the blends: structural mapping}
In Conceptual Blending theory, after the selection of input spaces, the subsequent step is to perform a partial matching between elements of the given mental spaces. This can be seen as establishing an analogy between the two inputs. 
	Our input spaces are in the form of semantic maps composed of $N_c$ concepts and $N_t$ triples, with $N_t, N_c \in \mathbb{N}$. The triples are in the form $<$$concept_0,relation,concept_1$$>$. Each concept corresponds to a vertex in a generic graph and the relation represents a directed edge connecting both concepts. 
	%Thus $<$$concept_0,relation,concept_1$$>$ corresponds to two vertices, $concept_0$ and $concept_1$, connected by the directed edge labeled as $relation$ from $concept_0$ to $concept_1$. Both concepts and relations are arrays of characters (strings).
	
	The \emph{Mapper} iterates through all possible root mappings, each composed of two distinct concepts taken from the input spaces. This means that there is a total of $\binom{N_c}{2}$ iterations. Then, the algorithm 
	%aims to find a structural isomorphism in the global input space, which 
	extracts two isomorphic sub-graphs from the larger input space. The two sub-graphs are split
	%composed from a split of the larger graph 
	in two sets of vertices $A$ (left) and $B$ (right). The structural isomorphism is defined by the sequence of relation types (pw, isa,...) found in both sub-graphs.
	
	Starting at the root mapping defined by two (left and right) concepts, the isomorphic sub-graphs are extracted from the larger semantic structure (the input spaces) by executing two synchronised expansions of nearby concepts at increasingly depths. The first expansion starts from the left concept and the second from the right concept. The left expansion is done recursively in the form of a depth first expansion and the right as a breadth first expansion.
	The synchronisation is controlled by two mechanisms:
	\begin{enumerate}
		\item the depth of the expansion, which is related to the number of relations reached by each expansion, starting at either concept from the root mapping;
		\item the label used for selecting the same relation to be expanded next in both sub-graphs.
	\end{enumerate}
	
	Both left (depth) and right (breadth) expansions are always synchronized at the same level of deepness (first mechanism above).
	
	While expanding, the algorithm stores additional associations between each matched relations and the corresponding concept which was reached through that relation. In reality, what is likely to happen is to occur a multitude of isomorphisms. In that case, the algorithm will store various mappings from any given concept to multiple different concepts, as long as the same concepts were reached from a previous concept with the same relation. In the end, each isomorphism and corresponding set of concept mappings gives rise to an analogy. The output of the \emph{Mapper} component is a list of analogies with the greatest number of mappings.

\subsection{Generating the blends: construction and relations}

The \emph{Visual Blender} component uses structured base-representations (of the input concepts) along with their set of relations among parts to produce visual blends based on analogies (mappings) produced by the \emph{Mapper} component.

The way of structuring the representations is based on the \emph{Syntactic decomposition of graphic representations} proposed by von Engelhardt \shortcite{engelhardt2002} in which a composite graphic object consists of: a graphic space (occupied by the object); a set of graphic objects (which may also be composite graphic objects); and a set of graphic relations (which may be \emph{object-to-space} and/or \emph{object-to-object}).

The objects store several attributes: name, shape, position relative to the father-object (which has the object in the set of graphic objects), the set of relations to other objects and the set of child-objects. By having such a structure, the complexity of blending two base representations is reduced, as it facilitates object exchange and recursive changing (by moving an object, the child-objects are also easily moved).

A relation between two objects consists of: the object \emph{A}, the object \emph{B} and the type of relation (\emph{above, lowerPart, inside, ...}) -- e.g. \emph{eye (A)} \underline{inside} \emph{head (B)}.

\subsection{Generating the blends: visual blending}

The \emph{Visual Blender} receives the analogies between two given concepts produced by the \emph{Mapper} component and the blend step occurs during the production of the visual representation -- differently from what happens in \emph{The Boat-House Visual Blending Experience} \cite{pereira2002boat}, in which the blends are merely interpreted at the visual representation level.

\begin{figure}
\includegraphics[width=8.41 cm]{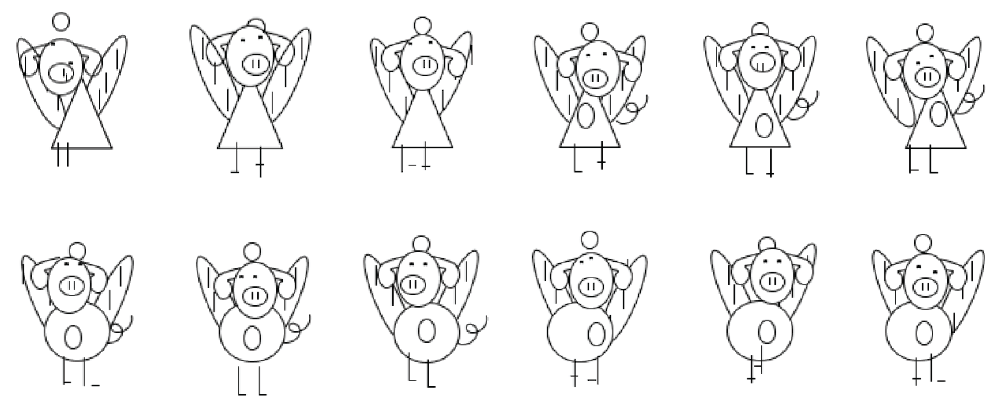}
\caption{The ``face expressions'' of the angel-pigs -- given the same or similar rules, the produced results are still quite diverse. The \emph{tail} and the \emph{belly / round shape} in the triangular \emph{body} are obtain through a process of composition (S4.4).
}
\label{fig:diversity} %id para referenciar
\end{figure}

The part of the blending process that occurs at the \emph{Visual Blender} produces visual representations as output and consists of five steps:
\begin{description}
\item [S1] An analogy is selected from the set of analogies provided by the \emph{Mapper};
\item [S2] One of the concepts (either \emph{A} or \emph{B}) is chosen as a base (consider \emph{A} as the chosen one, as an example); 
\item [S3] A visual representation (\emph{rA}) is chosen for the concept \emph{A} and a visual representation (\emph{rB}) is chosen for the concept \emph{B};
\item [S4] Parts of \emph{rA} are replaced by parts of \emph{rB} based on the analogy. For each mapping of the analogy -- consider for example \emph{leg} of \emph{A} corresponds to \emph{arm} of \emph{B} -- the following steps occur:
\begin{description}
\item [S4.1] The parts from \emph{rA} that correspond to the element in the mapping (e.g. \emph{leg}) are searched using the names of the objects. In the current example, the parts found could be \emph{left\_leg} (\emph{left\_} is a prefix), \emph{right\_leg\_1} (\emph{right\_} is a prefix and \emph{\_1} a suffix) or even \emph{leftfront\_leg};
\item [S4.2] For each of the found parts in S4.1, a matching part is searched in \emph{rB} using the names of the objects. This search firstly looks for objects that match the full name, including the prefix and suffix  (e.g. \emph{right\_arm\_1}) and, if none is found, searches only using the name in the mapping (e.g. \emph{arm}). It avoids plural objects (e.g. \emph{arms}). If no part is found, it proceeds to step S4.4;
\item [S4.3] The found part (\emph{pA}) of \emph{rA} is replaced by the matching part (\emph{pB}) of \emph{rB}, updating the relative positions of \emph{pB} and its child-objects, and relations (i.e. relations that used to belong to \emph{pA} now point to \emph{pB});
\item [S4.4] A process of Composition occurs (see examples in Fig.\ref{fig:diversity} -- the \emph{tail} and the \emph{belly / round shape} in the triangular \emph{body} are obtain using composition). For each of the matching parts from \emph{rB} (even if the replacement does not occur) a search is done for parts from \emph{rB} that have a relation with \emph{pB} (for example, a found part could be \emph{hand}). It only accepts a part if \emph{rA} does not have a part with the same name and if the analogy used does not have a mapping for it. If a found part matches these criteria, a composition can occur by copying the part to \emph{rA} (in our example, depending on either the replacement in Step S4.3 occurred or not, \emph{rA} would have either \emph{hand} related to \emph{arm} or to \emph{leg}, respectively);
\end{description}
\item [S5] The \emph{rA} resulting from the previous steps is checked for inconsistencies (both in terms of relative positioning and obsolete relations -- which can happen if an object does not exist anymore due to a replacement);
\end{description} 

After generating a representation, the similarity to the base representations (\emph{rA} and \emph{rB}) is assessed to avoid producing representations visually equal to them. This assessment is done by using a Root Mean Square Error (RMSE) measure that checks the similarity on a pixel-by-pixel basis.

\subsection{Evolutionary Engine}

The main goal of the \emph{Visual Blender} component is to produce and evolve possible visual blends based on the analogies produced by the \emph{Mapper}. In order to achieve this and promote diversity while respecting each analogy, an evolutionary engine was implemented. This engine is based on a Genetic Algorithm (GA) using several populations (each corresponding to a different analogy), in which each individual is a visual blend.

In order to guide evolution, we adopt a fitness function that assesses how well the the existing relations are respected. Some of the relations, e.g. the relation \emph{above},  have a binary assessment -- either 0, when the relation is not respected, or 1 when it is respected. Others yield a value between 0 and 1 depending on how respected it is -- e.g. the relation \emph{inside} calculates the number of points that are inside and returns $\frac{\#PointsInside}{total\#Points}$.

The fitness function for a given visual blend $b$ is as follows:

\begin{equation}
f(b) = \frac{\sum\limits^{\#R(b)}_{i=1}v(r_{i}(b))}{\#R(b)}, 
\end{equation}

\noindent where $\#R(b)$ denotes the number of relations present in $b$ and $v$ is the function with values in $[0,1]$ that indicates how much a relation $r$ is respected ($0$ -- not respected at all, 1 -- fully respected).  

The evolutionary engine includes five tasks which are performed in each generation for each population:
\begin{description}
\item [T1] Produce more individuals when the population size is below the maximum size;
\item [T2] Store the  best individual to avoid loosing it (elitism);
\item [T3] Mutate the individuals of the population. For each individual, each object can be mutated by changing its position. This change also affects its child-objects;
\item [T4] Recombine the individuals: the parents are chosen using tournament selection (with size 2) and a \emph{N-point crossover} is used to produce the children. In order to avoid the generation of invalid individuals, the crossover only occurs between chromosomes (objects) with the same name (e.g. a \emph{head} is only exchanged with a \emph{head}). If this rule was not used, it would lead to the production of descendants that would not respect the analogy followed by the population;
\item [T5] Removal of identical individuals in order to increase variability. 
\end{description}

In the experiments reported in this paper the mutation probability was set to $0.05$, per gene, and the recombination probability to $0.2$, per individual. These values were established empirically in preliminary runs.

\section{Results and discussion}

In this section we present and discuss the experimental results. We begin with a general analysis. Afterwards, we analyse the resulting visual representations comparing them with the data collected in the initial enquiry. Then, we analyse the quality of the produced blends by presenting the results of a final enquiry focused on perception.
%general results

\begin{figure}
\includegraphics[width=8.41 cm]{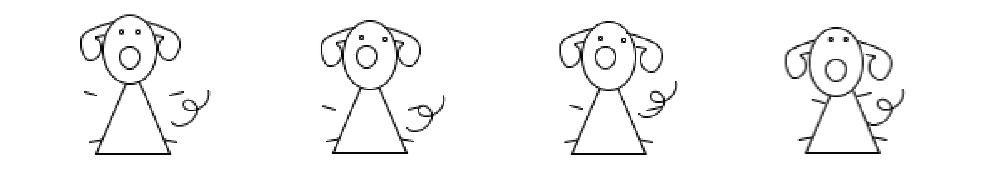}
\caption{Evolution of a blend: the \emph{legs} and \emph{tail} come closer to the \emph{body}, guided by the fitness function.
}
\label{fig:evolution} %id para referenciar
\end{figure}

Overall, the analysis of the experimental results indicates that the implemented blender is able to produce sets of blends with great variability (see Fig.\ref{fig:diversity} for an example of the results obtained for the same analogy and the same relations) and unexpected features, while respecting the analogy. The evolutionary engine is capable of evolving the blends towards a higher number of satisfied relations. This is verifiable in numerical terms, through the analysis of the evolution of fitness, and also through the visual assessment of the results. Figure \ref{fig:evolution} illustrates the evolution of a blend: the \emph{legs} and \emph{tail} are iteratively moved towards the \emph{body} in order to increase the degree of satisfaction of the relations.

%initialization bias -> initialisation bias

We can also observe that the system tends to produce blends in which few parts are exchanged between concepts. This can be explained as follows: when the number of parts increases the difficulty of (randomly) producing a blend with adequate fitness drastically decreases. As such, blends with fewer exchanges of parts, thus closer to base representation (in which all the relations are satisfied), tend to become dominant during the initial generations of the evolutionary runs. We consider that a significantly higher number of runs would be necessary to produce blends with more exchanges. Furthermore, valuing the exchange of parts, through the modification of the fitness function, may also be advisable for promoting the emergence of such blends.

%Partes criterios do blend.
% Nevertheless, it is also important to consider that this feature is in agreement with the Topology principle -- from the Optimality Principles presented by Fauconnier and Turner \shortcite{FauconnierTurner:98} -- which values the similarity degree to one of the input spaces (in this case, with one of the base representations).
As the blends are being produced as a visual representation which works as a whole as well as a set of individual parts, the Principle of Integration is being respected by design -- from the Optimality Principles presented by Fauconnier and Turner \shortcite{FauconnierTurner:98}.

\begin{figure}
\includegraphics[width=8.41 cm]{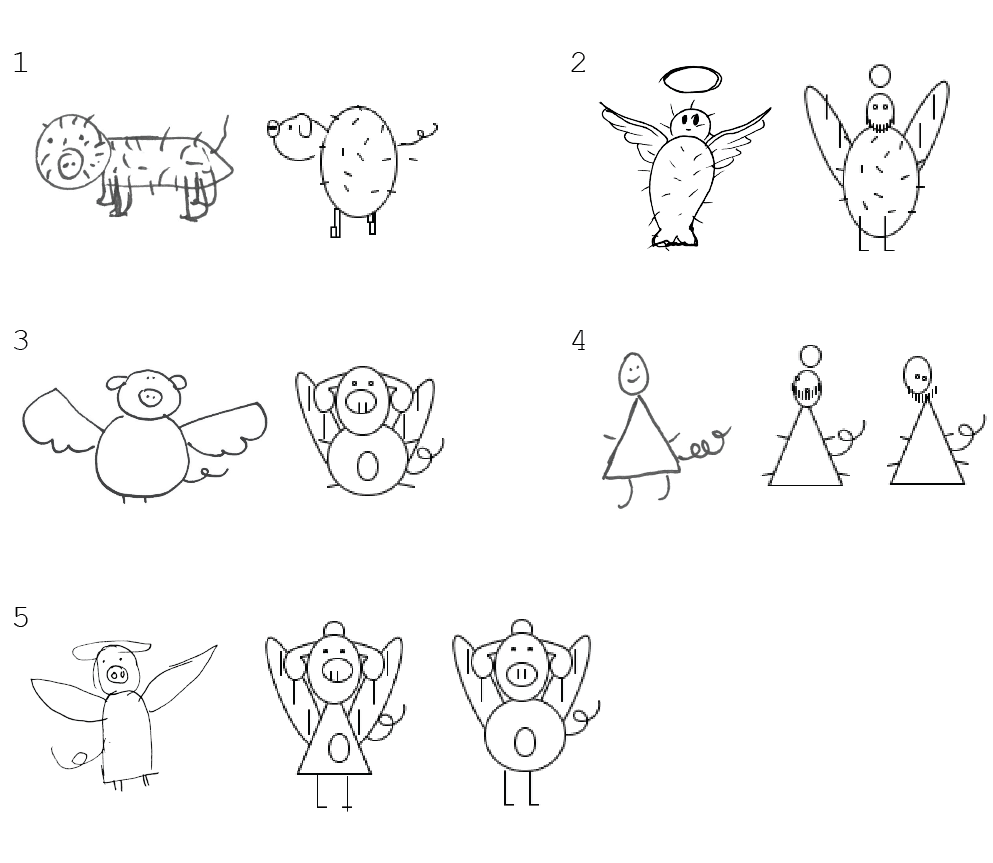}
\caption{Comparison between hand-drawn blends and blends generated by the implemented \emph{Blender}, organised by groups: group  1 corresponds to \emph{pig-cactus blends}; 2 corresponds to \emph{angel-cactus}; groups 3-5 correspond to \emph{pig-angel} (the figure on the left of each group is the hand-drawn blend).
}
\label{fig:blendComp} %id para referenciar
\end{figure}

\subsection{Comparison with user-drawn blends}
During the initial phase of the project, we conducted a task of collecting visual blends drawn by the participants. A total of 39 drawn blends were collected, from which 14 correspond to the blend between \emph{cactus} and \emph{angel}, 12 correspond to the blend between \emph{cactus} and \emph{pig} and 13 correspond to the blend between \emph{pig} and \emph{cactus}. 
The implemented blender was able to produce visual blends similar to the ones drawn by the participants (see some examples in Fig. \ref{fig:blendComp}). After analysing the produced blends, the following results were obtained:
\begin{itemize}
\item 23 from the 39 drawn blends (DB) were produced by our \emph{Blender};
\item 2 are not possible to be produced due to inconsistencies (e.g. one drawn blend from \emph{angel-pig} used a mapping from \emph{wing}-\emph{tail} and at the same time maintained the \emph{wings});
\item 6 were not able to be produced in the current version due to mappings that were not produced by the \emph{Mapper} (e.g. \emph{head} from \emph{angel} with \emph{body} from \emph{cactus});
\item 5 were not able to be produced because not all of the collected drawn representations were used in the experiments.
\end{itemize}

According to the aforementioned results, the implemented \emph{Blender} is not only able to produce blends that are coherent with the ones drawn by participants but is also able to produce novel blends that no participant drew, showing creative behaviour.

\begin{figure}
\includegraphics[width=8.41 cm]{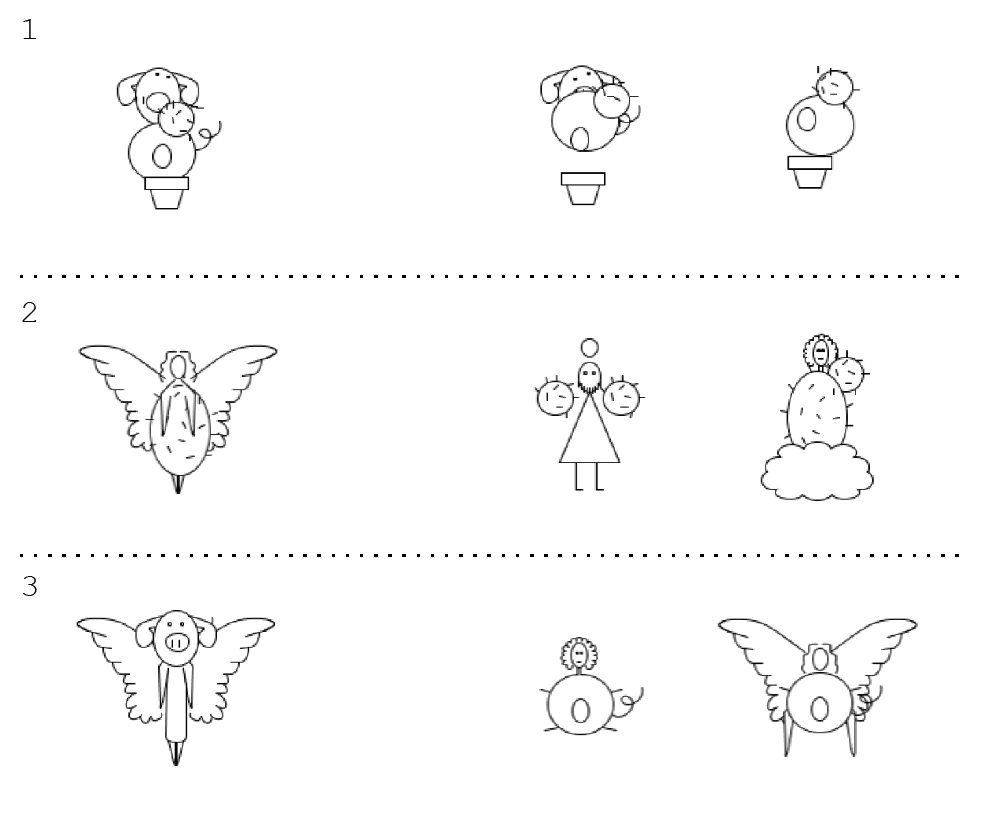}
\caption{Examples of the visual blends presented in the second enquiry. On the left are the ``good'' blends (one for each) and on the right are the ``bad'' blends (1 corresponds to \emph{cactus-pig}, 2  to \emph{angel-cactus} and 3 to \emph{angel-pig}).
}
\label{fig:goodBad} %id para referenciar
\end{figure}

\subsection{Evaluating perception} % Perceptual evaluation? (amilcar)

In order to assess if the produced blends could be correctly perceived, a second enquiry was conducted. The main goal was to evaluate whether or not the participant could identify the input spaces used for each blend (i.e. if it was possible to identify \emph{pig} and \emph{cactus} in a blend produced for \emph{pig-cactus}). This is related to the Unpacking Principle \cite{FauconnierTurner:98}.

In the first enquiry, the fourth task (T4) consisted in collecting the prototypical parts for each concept -- these are the parts that most identify the concept (e.g. \emph{wing} for \emph{angel}).  We used the data collected for producing the second enquiry. For each blend (\emph{angel-pig}, \emph{cactus-pig} or \emph{angel-cactus}), four visual blends were selected (two considered ``good'' and two considered ``bad'', see Fig. \ref{fig:goodBad}). The quality evaluation (``bad'' or ``good'') was based on two criteria: fitness of the individual and presence or legibility of the prototypical parts (i.e. a ``good'' exemplar is an individual with the prototypical parts clearly identifiable; a ``bad'' exemplar is an individual with fewer prototypical parts or these are not clearly identifiable).

\begin{table}[h]
\caption{Number of correct names given (input spaces' names) for each of the blends (percentage of answers).}\label{tab:tab1}
\begin{center}
\begin{tabular}{lllll}
\toprule
                              &      & 0  & 1  & 2  \\ \midrule
\multirow{2}{*}{cactus-pig}   & Good & 20 & 50 & 30 \\
                              & Bad  & 50 & 50 & 0  \\
\midrule
\multirow{2}{*}{angel-pig}    & Good & 10 & 20 & 70 \\
                              & Bad  & 40 & 50 & 10 \\
\midrule
\multirow{2}{*}{angel-cactus} & Good & 0  & 60 & 40 \\
                              & Bad  & 10 & 80 & 10 \\ \bottomrule 
\end{tabular}
\end{center}
\end{table}

\begin{table}[h]
\centering
\caption{Number of correct names given (input spaces' names) for each of the blends (number of answers).}
\label{tab:tab2}
\begin{tabular}{@{}cccccc@{}}
\toprule
                              &                       & \# R. & 0 & 1 & \multicolumn{1}{l}{2} \\ \midrule
\multirow{4}{*}{cactus-pig}   & \multirow{2}{*}{Good} & 1    & 1 & 2 & 2                     \\
                              &                       & 2    & 1 & 3 & 1                     \\
                              & \multirow{2}{*}{Bad}  & 3    & 1 & 4 & 0                     \\
                              &                       & 4    & 4 & 1 & 0                     \\
\midrule
\multirow{4}{*}{angel-pig}    & \multirow{2}{*}{Good} & 5    & 0 & 1 & 4                     \\
                              &                       & 6    & 1 & 1 & 3                     \\
                              & \multirow{2}{*}{Bad}  & 7    & 2 & 3 & 0                     \\
                              &                       & 8    & 2 & 2 & 1                     \\
\midrule
\multirow{4}{*}{angel-cactus} & \multirow{2}{*}{Good} & 9    & 0 & 4 & 1                     \\
                              &                       & 10   & 0 & 2 & 3                     \\
                              & \multirow{2}{*}{Bad}  & 11   & 0 & 5 & 0                     \\
                              &                       & 12   & 1 & 3 & 1                     \\ \bottomrule 
\end{tabular}
\end{table}

A total of 12 visual blends were used and the enquiry was conducted to 30 participants. Each visual blend was tested by 5 participants. In order minimise the biasing of the results, each participant evaluated two visual representations (one ``bad'' and one ``good'') of different blends (e.g. when the first was of \emph{cactus-pig}, the second could only be of \emph{angel-pig} or \emph{angel-cactus}). The ``bad'' blends were evaluated first to further minimise the biasing.

The results (Table \ref{tab:tab1} and Table \ref{tab:tab2}), clearly show that the ``good'' blends were easier to be correctly named (the percentage of total correct naming is always higher for the ``good'' examples; the percentage of total incorrect naming is always higher for the ``bad'' blends). In addition to this, the names of the input spaces were also easier to be identified in some of the representations than in others (e.g. the ``good'' blends for \emph{angel-pig} received more totally correct answers than the rest of the blends, as shown in Table~\ref{tab:tab2}).

Overall, the majority of the participants could identify at least one of the input spaces for the ``good'' exemplars of visual blends. Even though some of the participants could not correctly name both of the input spaces, the answers given were somehow related to the correct ones (e.g. the names given for the input spaces in  the first ``bad'' blend of 3 in Fig. \ref{fig:goodBad} were often \emph{pig} and \emph{lady/woman}, instead of \emph{pig} and \emph{angel} -- this is due to the fact that no \emph{halo} nor \emph{wings} are presented).

\section{Conclusions and future work}

We presented a descriptive approach for automatic generation of visual blends. The approach uses structured representations along with sets of visual relations which describe how the parts -- in which the visual representation can be decomposed -- relate among each other. The experimental results demonstrate the ability of the \emph{Blender} to produce analogies from input mental spaces and generate a wide variety of visual blends based on them. The \emph{Visual Blender} component, in addition to fulfilling its purpose, is able to produce interesting and unexpected blends. 
Future enhancements to the proposed approach include:
\begin{description}
\item [(i)] exploring an island approach in which exchange of individuals from different analogies may occur if they respect the analogy of the destination population; 
\item [(ii)] exploring the role of the user (guided evolution), by allowing the selection of individuals to evolve;
\item [(iii)] considering Optimality Principles in the assessment of fitness (e.g. how many parts are exchanged) and exploring which of them may be useful or needed -- something discussed by Martins et al. \shortcite{martins2016optimality};
\item [(iv)] using relations such as \emph{biggerThan} or \emph{smallerThan} to explore style changing  (e.g. the style of the produced blends will be affected if a base visual representation has \emph{head} \underline{biggerThan} \emph{body}); 
\item [(v)] exploring context in the production of blends (e.g. stars surrounding the angel).
\end{description}

%future stuff and advantages
% Questao do estilo(TIAGO)
% Questao do contexto (FILIPE)
%explor on fitness criteria blending (how many parts are exchanged and how many relations are respected)
%Voltar a referir hybrid blend generation 
%futuro: declarativa e procedimental
%ser capaz de calcular a relevancia
%híbrido: blend dos dois lados
%Why blender? Why better? (elaboration, easiness?…)

\section{Acknowledgements}

This research is partially funded by: Funda\c{c}\~{a}o para a Ci\^{e}ncia e Tecnologia (FCT), Portugal, under the grant SFRH/BD/120905/2016.

\bibliographystyle{iccc}
\bibliography{coimbra}

\end{document}